\documentclass[copyright,creativecommons]{eptcs}
\providecommand{\event}{} 
\providecommand{\volume}{2}
\providecommand{\anno}{2009} 
\providecommand{\firstpage}{51}
\usepackage{breakurl}

\usepackage{amsmath, amssymb, amsxtra, amsfonts}

\usepackage{graphicx}

\newcommand{\s}[1]{\mbox{\sf{#1}}}
\newcommand{\so}[1]{\mbox{$\overline{\s{#1}}$}}

\def\odiv{{ \ominus\hspace{-6.7pt} \div}}
\def\ent{\vdash}

\def\0{{\bf 0}}
\def\1{{\bf 1}}
\newtheorem{thm}{Theorem}[section]
\newtheorem{definition}[thm]{Definition}
\newtheorem{example}[thm]{Example}

\newcommand{\la}{\langle}
\newcommand{\ra}{\rangle}

\newcommand{\rrarrow}{\longrightarrow}

\newcommand{\tell}{{\bf tell}}

\newcommand{\ask}{{\bf ask}}

\def\0{{\mathbf 0}}
\def\1{{\mathbf 1}}
\def\C{{\mathcal C}}

\def\titlerunning{Soft Constraints for Quality Aspects in Service Oriented Architectures}
\def\authorrunning{S.~Bistarelli \& F.~Santini}

\begin{document}

\title{Soft Constraints for Quality Aspects in Service Oriented Architectures}

\author{Stefano Bistarelli
\institute{Dipartimento di Matematica Informatica\\ Universit\`a di
Perugia, Italy} \institute{Istituto di Informatica e Telematica
(CNR)\\ Pisa, Italy} \institute{Dipartimento di Scienze,
Universit\`a ``G. d'Annunzio''\\ Chieti-Pescara, Italy}
\email{bista@dipmat.unipg.it}\\ {\footnotesize\tt stefano.bistarelli@iit.cnr.it}\\ {\footnotesize\tt bista@sci.unich.it}
\and Francesco Santini \institute{Dipartimento di Scienze,
Universit\`a ``G. d'Annunzio''\\ Chieti-Pescara, Italy}
\institute{Istituto di Informatica e Telematica (CNR)\\ Pisa,
Italy} \email{santini@sci.unich.it}\\ {\footnotesize\tt francesco.santini@iit.cnr.it}
}

\maketitle

\begin{abstract}
We propose the use of Soft Constraints as a natural way to model
Service Oriented Architecture. In the framework, constraints are
used to model components and connectors and constraint aggregation
is used to represent their interactions. The ``quality of a
service'' is measured and considered when performing queries to
service providers. Some examples consist in the levels of cost,
performance and availability required by clients. In our
framework, the QoS scores are represented by the softness level of
the constraint and the measure of complex (web) services is
computed by combining the levels of the components.
\end{abstract}

\section{Introduction}\label{Introduction}

Constraint programming is a powerful paradigm for solving
combinatorial search problems that draws on a wide range of
techniques from artificial intelligence, computer science,
databases, programming languages, and operations
research~\cite{sumconstraints,bistabook,bookrossi}. It is
currently applied with success to many domains, such as
scheduling, planning, vehicle routing, configuration, networks,
and bioinformatics. The basic idea in constraint programming is
that the user states the constraints and a general purpose
constraint solver solves them. Constraints are just relations, and
a \emph{Constraint Satisfaction Problem} (CSP) states which
relations should hold among the given decision variables (we refer
to this classical view as ``crisp'' constraints). Constraint
solvers take a real-world problem, represented in terms of
decision variables and constraints, and find an assignment of
values to all the variables that satisfies all the constraints.

Rather than trying to satisfy a set of constraints, sometimes
people want to optimize them. This means that there is an
objective function that tells us the quality of each solution, and
the aim is to find a solution with optimal quality. For example,
fuzzy constraints~\cite{sumconstraints,bistabook,bookrossi} allow
for the whole range of satisfiability levels between $0$ and $1$.
In weighted constraints, instead, each constraint is given a
weight, and the aim is to find a solution for which the sum of the
weights of the satisfied constraints is maximal.

The idea of the semiring-based
formalism~\cite{sumconstraints,bistabook} was to further extend
the classical constraint notion, and to do it with a formalism
that could encompass most of the existing extensions, as well as
other ones not yet defined, with the aim to provide a single
environment where properties could be proven once and for all, and
inherited by all the instances. At the technical level, this was
done by adding to the usual notion of a CSP the concept of a
structure representing the levels of satisfiability of the
constraints. Such a structure is a set with two operations (see
Sec.~\ref{sec:bgcon} for further details): one (written $+$) is
used to generate an ordering over the levels, while the other one
($\times$) is used to define how two levels can be combined and
which level is the result of such combination. Because of the
properties required on such operations, this structure is similar
to a semiring (see Sec.~\ref{sec:bgcon}): from here the
terminology of ``semiring-based soft
constraint''~\cite{sumconstraints,bistabook} (and
Sec.~\ref{sec:bgcon}), that is, constraints with several levels of
satisfiability, and whose levels are (totally or partially)
ordered according to the semiring structure. In general, problems
defined according to the semiring-based framework are called
\emph{Soft Constraint Satisfaction Problems} (SCSPs).

The aim of this paper is to apply  \emph{Quality of Service} (QoS)
measures for \emph{Service Oriented Architectures}
(SOAs)~\cite{soa2,soa}. Such architecture outlines a way of
reorganizing software applications and infrastructure into a set
of interacting services and aims at a loose coupling of services
with operating systems, programming languages and other
technologies. A SOA separates functions into distinct units or
services, and these services communicate with each other by
passing data from one service to another, or by coordinating an
activity between two or more services. \emph{Web
services}~\cite{wslaframework,federated} can implement a
service-oriented architecture.

SOAs clearly represent a distributed environment and QoS aspects
become very important to evaluate, since the final integrated
service must fulfill the non-functional requirements of the final
users; this composition needs to be monitored~\cite{soa2,soa}. We
are also interested in representing contracts and \emph{Service
Level Agreements}~\cite{federated,wslaframework} (SLAs) in terms
of constraint based languages. The notions of contract and SLAs
are very important in SOC since they allow to describe the mutual
interactions between communicating parties and to express
properties related to the quality of service such as cost,
performance, reliability and availability. The existing languages
for describing Web services (e.g.~\emph{WSDL}, \emph{WS-CDL} and
\emph{WS-BPEL}) are not adequate for describing contracts and SLAs
and, so far, there exists no agreement on a specific proposal in
this sense: a general, established theory of contracts is still
missing~\cite{federated,wslaframework}.

The key idea of this paper is to use the a soft constraint
framework in order to be able to manage SOAs in a declarative
fashion by considering together both the requirements/interfaces
of each service and their QoS
estimation~\cite{wirsing,doctoral1,doctoral2}. C-semirings can
represent several  QoS attributes, while soft constraints
represent the specification of each service to integrate: they
link these measures to the resources spent in providing it, for
instance, ``the reliability is equal to 80\% plus 5\% for each
other processor used to execute the service''. This statement can
be easily represented with a soft constraint where the number of
processors corresponds to the $x$ variable, and the preference
(i.e.~reliability) level is given by the $5x + 80$ polynomial.

Beside expressivity reasons, other advantages w.r.t. crisp
constraints  are that soft constraints can solve over-constrained
problems (i.e.~when it is not possible to solve all of them at the
same time) and that, when we have to deal with quality, many
related concepts are ``smooth'': quality can be represented with
intervals of ``more or less'' acceptable values. It has been
proved that constraint in general are a powerful paradigm for
solving combinatorial search
problems~\cite{sumconstraints,bistabook,bookrossi}. Moreover,
there exists a wide body of existing research results on solving
(soft) CSP for large systems of constraints in a fully mechanized
manner~\cite{sumconstraints,bistabook}.

The paper is organized as follows: Sec.~\ref{sec:bgcon} presents
the minimum background notions needed to understand soft
constraints, while Sec.~\ref{sec:Dependqual} closes the
introductory part by defining SOAs, QoS aspects and by showing how
semiring instantiations can represent these non-functional
aspects. Sec.~\ref{sec:conn} shows that the use of soft
constraints permits us to perform a quantitative analysis of
system integrity. Section~\ref{sec:nego} shows how QoS can be
modeled and
checked by using a soft constraint-based formal language. 
Finally,
Sec.~\ref{sec:related} present the related work, while
Sec.~\ref{sec:onclusions} draws the final conclusions and
discusses the directions for future work.

\section{Background on Soft Constraints}\label{sec:bgcon}

\paragraph{Absorptive Semiring.}\label{sec:division}

An absorptive semiring~\cite{ecai06} $S$ can be represented as a
$\langle A,+,\times,\0,\1 \rangle$ tuple such that: \textit{i)}
$A$ is a set and $\0, \1 \in A$; \textit{ii)} $+$ is commutative,
associative and $\0$ is its unit element; \textit{iii)} $\times$
is associative, distributes over $+$, $\1$ is its unit element and
$\0$ is its absorbing element. Moreover, $+$ is idempotent, $\1$
is its absorbing element and $\times$ is commutative.
Let us consider the relation $\leq_S$ over $A$ such that $a \leq_S
b$ iff $a+b = b$. Then it is possible to prove that
(see~\cite{jacm}): \textit{i)} $\leq_S$ is a partial order;
\textit{ii)} $+$ and $\times$ are monotonic on $\leq_S$;
\textit{iii)} $\0$ is its minimum and $\1$ its maximum;
\textit{iv)} $\langle A,\leq_S \rangle$ is a complete lattice and,
for all $a, b \in A$, $a+b = lub(a,b)$ (where $lub$ is the {\em
least upper bound}). Informally, the relation $\leq_S$ gives us a
way to compare semiring values and constraints. In fact, when we
have $a \leq_S b$ (or simply $a \leq b$ when the semiring will be
clear from the context), we will say that {\em b is better than
a}.

In~\cite{ecai06} the authors extended the semiring structure by
adding the notion of \emph{division}, i.e.~$\div$, as a weak
inverse operation of $\times$. An absorptive semiring $S$ is
\emph{invertible} if, for all the elements $a, b \in A$ such that
$a\leq b$, there exists an element $c \in A$
such that $b \times c = a$~\cite{ecai06}. 
If $S$ is absorptive and invertible, then, $S$ is \emph{invertible
by residuation} if the set $\{x \in A \mid b \times x = a\}$
admits a maximum for all elements $a, b \in A$ such that $a \leq
b$~\cite{ecai06}. Moreover, if $S$ is absorptive, then it is
\emph{residuated} if the set $\{x \in A \mid b \times x \leq a\}$
admits a maximum for all elements $a, b \in A$, denoted $a \div
b$. With an abuse of notation, the maximal element among solutions
is denoted $a\div b$. This choice is not ambiguous: if an
absorptive semiring is invertible and residuated, then it is also
invertible by residuation, and the two definitions yield the same
value.

To use these properties, in~\cite{ecai06} it is stated that if we
have an absorptive and complete semiring\footnote{If $S$ is an
absorptive semiring, then $S$ is complete if it is closed with
respect to infinite sums, and the distributivity law holds also
for an infinite number of summands.}, then it is residuated.  For
this reason, since all classical soft constraint instances (i.e.~\emph{Classical CSPs}, \emph{Fuzzy CSPs}, \emph{Probabilistic
CSPs} and \emph{Weighted CSPs}) are complete and consequently
residuated, the notion of semiring division (i.e.~$\div$) can be
applied to all of them. 

\paragraph{Soft Constraint System.}
A {\em soft constraint}~\cite{jacm,bistabook} may be seen as a
constraint where each instantiation of its variables has an
associated preference. Given  $S = \langle A,+,\times,\0,\1
\rangle$ and an ordered set of variables $V$ over a finite domain
$D$, a soft constraint is a function which, given an assignment
$\eta : V\rightarrow D$ of the variables, returns a value of the
semiring.
Using this notation  
$\C = \eta \rightarrow A$ is the set of all possible constraints
that can be built starting from $S$, $D$ and $V$.

Any function in $\C$ involves all the variables in $V$, but we
impose that it depends on the assignment of only a finite subset
of them. So, for instance, a binary constraint $c_{x,y}$ over
variables $x$ and $y$, is a function $c_{x,y}: (V\rightarrow
D)\rightarrow A$, but it depends only on the assignment of
variables $\{x,y\}\subseteq V$ (the {\em support} of the
constraint, or {\em scope}). Note that $c\eta[v:=d_1]$ means
$c\eta'$ where $\eta'$ is $\eta$
modified with the assignment $v:=d_1$. 
Notice also that, with $c\eta$, the result we obtain is a semiring
value, i.e.~$c\eta=a$.

Given set $\C$, the combination function $\otimes: \C\times\C
\rightarrow \C$ is defined as $(c_1\otimes c_2)\eta =
c_1\eta\times c_2\eta$ (see also \cite{jacm,bistabook,scc}).
Having defined the operation $\div$ on semirings, the constraint
division function $\odiv: \C\times\C \rightarrow \C$ is instead
defined as $(c_1\, \odiv \, c_2)\eta = c_1\eta\div
c_2\eta$~\cite{ecai06}. Informally, performing the $\otimes$ or
the $\odiv$ between two constraints means building a new
constraint whose support involves all the variables of the
original ones, and which associates with each tuple of domain
values for such variables a semiring element which is obtained by
multiplying or, respectively, dividing the elements associated by
the original constraints to the appropriate sub-tuples. The
partial order $\leq_S$ over $\C$ can be easily extended among
constraints by defining $c_1 \sqsubseteq c_2 \iff c_1 \eta \leq
c_2 \eta$. Consider set $\C$ and partial order
$\sqsubseteq$. Then an entailment relation $\ent \subseteq \wp(\C)
\times \C$ is defined s.t. for each $C \in \wp(\C)$ and $c \in
\C$, we have $C \ent c \iff \bigotimes C \sqsubseteq c$ (see
also~\cite{bistabook,scc}).

Given a constraint $c \in \C$ and a variable $v \in V$, the {\em
projection}~\cite{jacm,bistabook,scc} of $c$ over
$V\backslash\{v\}$, written $c\Downarrow_{(V \backslash\{v\})}$ is
the constraint $c'$ s.t. $c'\eta = \sum_{d \in D} c \eta [v:=d]$.
Informally, projecting means eliminating some variables from the
support. This is done by associating with each tuple over the
remaining variables a semiring element which is the sum of the
elements associated by the original constraint to all the
extensions of this tuple over the eliminated variables. To treat
the hiding operator of the language, a general notion of
existential quantifier is introduced by using notions similar to
those used in cylindric algebras. 
For each $x \in V$, the hiding function~\cite{bistabook,scc} is
defined as $(\exists_x c)\eta =\sum_{d_i\in D} c\eta[x := d_i]$.

To model parameter passing, for each $x,y \in V$ a diagonal
constraint~\cite{bistabook,scc} is defined as $d_{xy} \in \C$
s.t., $d_{xy}\eta[x:= a, y := b]= \1$ if $a=b$ and $d_{xy}\eta[x:=
a, y := b] = \0$ if $a\neq b$. Considering a semiring $S = \langle
A,+,\times,\0,\1 \rangle$, a domain of the variables $D$, an
ordered set of variables $V$ and the corresponding structure $\C$,
then $S_C=\langle \C, \otimes,\bar{\0},\bar{\1} , \exists_x,
d_{xy}\rangle$\footnote{$\bar{\0}$ and $\bar{\1}$ respectively
represent the constraints associating $\0$ and $\1$ to all
assignments of domain values;  in general, the $\bar{a}$ function
returns the semiring value $a$.} is a cylindric constraint system
({\em ``a la Saraswat''}~\cite{scc}).

\paragraph{Soft CSP and an Example.}A {\itshape Soft Constraint Satisfaction Problem}
(SCSP)~\cite{bistabook} defined as $P= \langle C, con \rangle$:
$C$ is the set of constraints and $con\subseteq V$ is the set of
variables of interest for the constraint set $C$, which however
may concern also variables not in $con$. This is called the {\em
best level of consistency} and it is defined by $blevel(P) =
Sol(P) \Downarrow_{\emptyset}$, where $Sol(P)=(\bigotimes
C)\Downarrow_{con}$; notice that $supp(blevel(P))= \emptyset$. We
also say that: $P$ is $\alpha$-consistent if $blevel(P) = \alpha$;
$P$ is consistent iff there exists $\alpha >_S \0$ such that $P$
is $\alpha$-consistent; $P$ is inconsistent if it is not
consistent.

\begin{figure}
\centering
\includegraphics[scale=0.65]{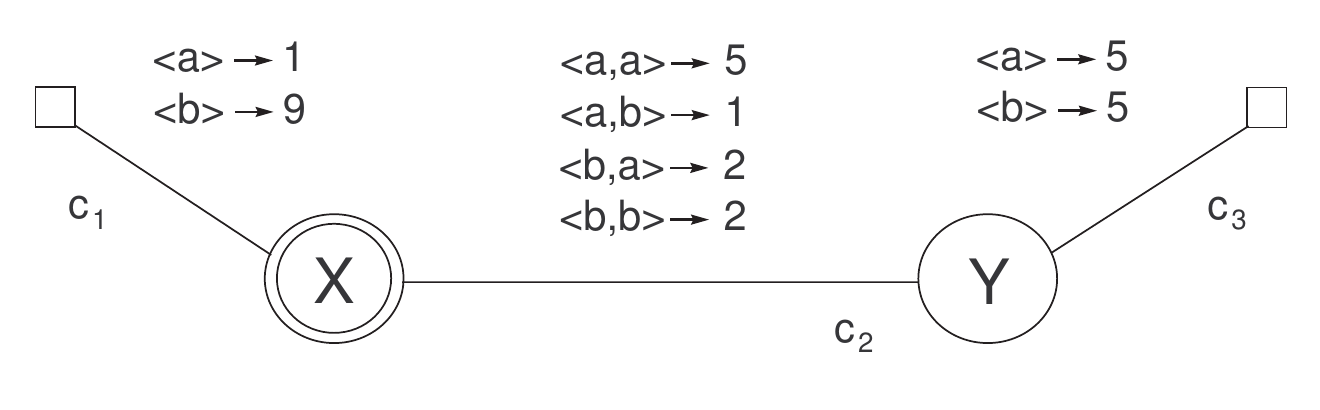} 
\caption{A soft CSP based on a Weighted semiring.}
\label{figure:wexample}
\end{figure}

Figure~\ref{figure:wexample} shows a weighted CSP as a graph.
Variables and constraints are represented respectively by nodes
and by undirected arcs (unary for $c_1$ and $c_3$, and binary for
$c_2$), and semiring values are written to the right of each
tuple. The variables of interest (that is the set $con$) are
represented with a double circle (i.e.~variable $X$). Here we
assume that the domain of the variables contains only elements $a$
and $b$. For example, the solution of the weighted CSP of
Fig.~\ref{figure:wexample} associates a semiring element to every
domain value of variable $X$. Such an element is obtained by first
combining all the constraints together. For instance, for the
tuple $\langle a, a\rangle$ (that is, $X = Y = a$), we have to
compute the sum of $1$ (which is the value assigned to $X = a$ in
constraint $c_1$), $5$ (which is the value assigned to $\langle X
= a, Y = a \rangle$ in $c_2$) and $5$ (which is the value for $Y =
a$ in $c_3$). Hence, the resulting value for this tuple is $11$.
We can do the same work for tuple $\langle a, b\rangle \rightarrow
7$, $\langle b, a\rangle \rightarrow 16$ and $\langle b, b\rangle
\rightarrow 16$. The obtained tuples are then projected over
variable x, obtaining the solution $\langle a \rangle \rightarrow
7$ and $\langle b \rangle \rightarrow 16$. The \emph{blevel} for
the example in Fig.~\ref{figure:wexample} is $7$ (related to the
solution $X = a$, $Y = b$).

\section{Service Oriented Architectures and QoS Aspects}\label{sec:Dependqual}

Service Oriented Architecture (SOA) can be defined as a group of
services, which communicate with each other~\cite{soa2,soa}. The
process of communication involves either simple data passing or it
could involve two or more services coordinating some activity.
Basic services, their descriptions, and basic operations
(publication, discovery, selection, and binding) that produce or
utilize such descriptions constitute the SOA foundation. The main
part of SOA is loose coupling of the components for integration.
Services are defined by their interface, describing both
functional and non-functional behaviour. Functional includes
describing data formats, pre and post conditions and the operation
performed by the service. Non-functional behaviour includes
security and other QoS parameters. The main four features of SOA
consist in \emph{Coordination}, \emph{Monitoring},
\emph{Conformance} and \emph{Quality of Service} (QoS)
composition~\cite{soa2}.

Services are self describing, open components that support rapid,
low-cost composition of distributed applications. Services are
offered by service providers, which are organizations that procure
the service implementations, supply their service descriptions,
and provide related technical and business support. Since services
may be offered by different enterprises and communicate over the
Internet, they provide a distributed computing infrastructure for
both intra and cross-enterprise~\cite{federated} application
integration and collaboration. Service descriptions are used to
advertise the service capabilities, interface, behaviour, and
quality. Publication of such information, about available
services, provides the necessary means for discovery, selection,
binding, and composition of services. Service clients (end-user
organizations that use some service) and service aggregators
(organizations that consolidate multiple services into a new,
single service offering) utilize service descriptions to achieve
their objectives.

QoS measures include also \emph{dependability} aspects:
dependability as applied to a computer system is defined by the
IFIP 10.4 Working Group on Dependable Computing and Fault
Tolerance as~\cite{ifip}: ``\emph{[..] the trustworthiness of a
computing system which allows reliance to be justifiably placed on
the service it delivers [..]}''.

Some different QoS/dependability measurements can be applied to a
system to determine its overall quality. A very general list of
attributes is: \emph{i)} \emph{Availability} - the probability
that a service is present and ready for use; \emph{ii)}
\emph{Reliability} - the capability of maintaining the service and
service quality; \emph{iii)} \emph{Safety} - the absence of
catastrophic consequences; \emph{iv)} \emph{Confidentiality} -
information is accessible only to those authorized to use it;
\emph{v)} \emph{Integrity} - the absence of improper system
alterations; and \emph{vi)} \emph{Maintainability} - to undergo
modifications and repairs. Some of these attributes, as
availability, are quantifiable by direct measurements (i.e.~they
are rather objective scores), but others are more subjective, e.g.~safety.

The semiring algebraic structures (see Sec.~\ref{sec:bgcon}) prove
to be an appropriate and very expressive cost model to represent
the QoS metrics shown in this Section. The cartesian product of
multiple c-semirings is still a c-semiring~\cite{bistabook} and,
therefore, we can model also a multicriteria optimization. In the
following list we present some possible semiring instantiations
and some of the possible metrics they can represent:

\begin{itemize}
    \item Weighted semirings $\langle \mathbb{R}^+, min, \hat{+}, \infty, 0
    \rangle$ ($\hat{+}$ is the arithmetic sum). In general, this semiring can
    represent \emph{additive} metrics: it can be used to count
    events or quantities to minimize the resulting sum,
    e.g.\,to save money while composing different services with different costs,
    or to minimize the downtime of service components (availability and reliability can be modeled this way).
    \item Fuzzy semirings $\langle [0,1], max, min, 0, 1 \rangle$.
    It can be used to represent fuzzy preferences on components, e.g.~\emph{low}, \emph{medium} or \emph{high} reliability when detailed information is not available.
    This semiring can be used to represent  \emph{concave}
    metrics, in which the composition result of all the values is obtained by ``flattening'' to the ``worst'' or ``best''
    value. One more application example is represented by the
    aggregation of bandwidth values along a network route or, however, by aggregating concave values on a pipeline of sub-services.
    \item Probabilistic semirings $\langle [0,1], max, \times, 0, 1
    \rangle$ ($\hat{\times}$ is the arithmetic multiplication). \emph{Multiplicative} me\-trics can be modeled with this
    semiring. As an example, this semiring can
    optimize (i.e.~maximize) the probability of successful behavior of services, by choosing the composition that
    optimizes the multiplication of probabilities. For example,
    the frequency of system faults can be studied from a
    probabilistic point of view; also availability can be represented with a percentage value.
    \item Set-Based semirings $\langle \mathcal{P}(A), \cup , \cap, \emptyset, A
    \rangle$. Properties and features of the service components can be
    represented with this semiring. For example, in order to
    represent related security rights, or time slots
    in which the services can be used (security issues).
    \item Classical semirings  $\langle \{0, 1\}, \vee , \wedge, 0, 1
    \rangle$. The classical semiring can be adopted to cast crisp
    constraints in the semiring-based framework defined
    in~\cite{bistabook,jacm97}. Even this semiring can be used
    to check if some properties are entailed by a service definition (i.e.~\emph{true} of \emph{false} values), by
    composing the properties of its components together.
\end{itemize}

\section{Soft Constraints to Enforce System Integrity}\label{sec:conn}
In this Section we show that soft constraints can model the
implementation of a service described with a policy
document~\cite{federated,wslaframework}; this really happens in
practice by using the \emph{Web Services Description Language}
(WSDL) that is an XML-based language that provides a model for
describing web services~\cite{wslaframework}. Moreover, by using
the projection operator (i.e.~$\Downarrow$ in
Sec.~\ref{sec:bgcon}) on this policy, which consists in the
composition (i.e.~$\otimes$ in Sec.~\ref{sec:bgcon}) of different
soft constraints, we obtain the external interface of the service
that are used to match the requests. This view can be used to
check the integrity of the system, that is if a particular service
ensures the consistency of actions, values, methods, measures and
principles; as a reminder, integrity can be seen as one of the QoS
attributes proposed in Sec.~\ref{sec:Dependqual}. The integrity
attribute is very important when different sub-services from
distinct providers are composed together to offer a single
structured service. The results presented here are inspired by the
work in~\cite{simon}.

\begin{figure}
\centering
    \includegraphics[scale=0.7]{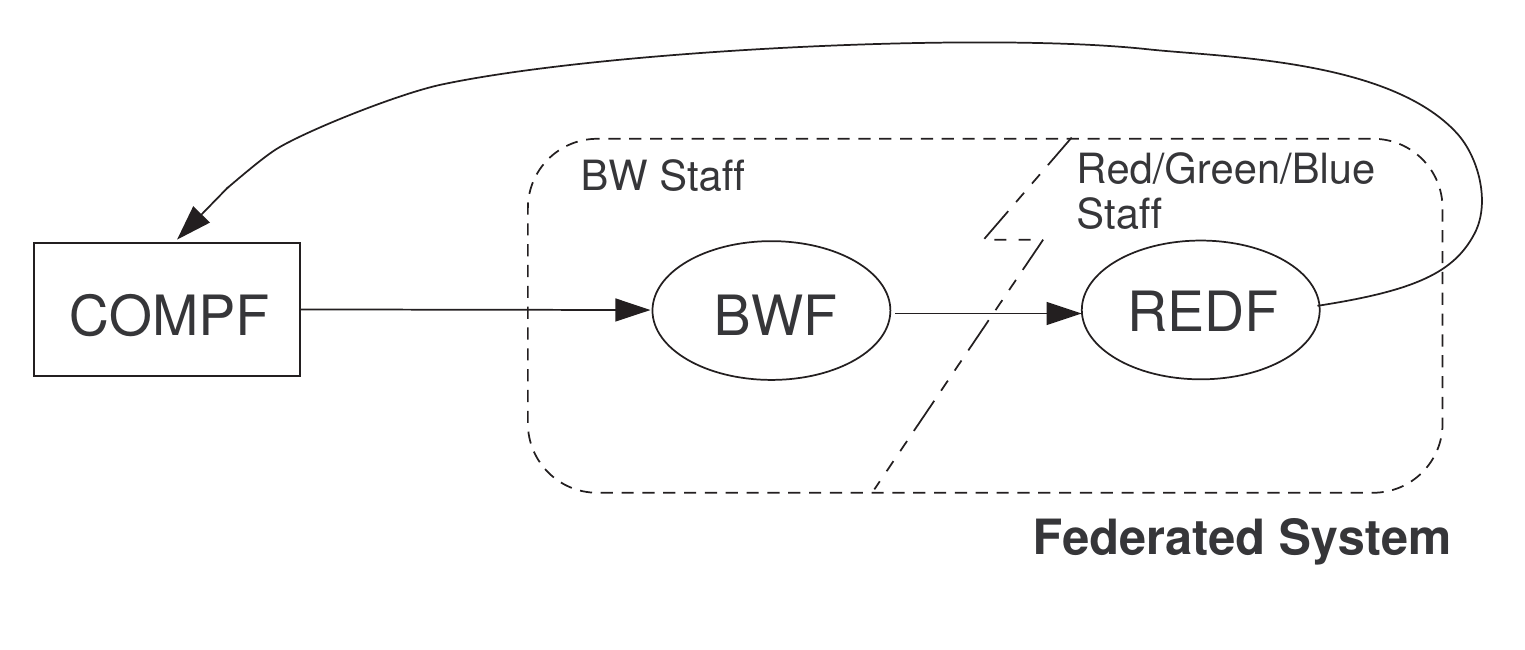}
\caption{A federated photo editing system.}\label{fig:federated}
\end{figure}

For the scenario example in Fig.~\ref{fig:federated}, 
suppose to have a digital photo editing service decomposed as a
set of sub-services; the compression/decompression module (i.e.~\emph{COMPF}) is located on the client side, while the other
filter modules are located on the side of the editing company and
can be reached through the network. The first module, i.e.~\emph{BWF} turns the colors in grey scale and the \emph{REDF}
filter absorbs green and blue and let only red become lighter. The
client wants to compress (e.g.~in a \emph{JPEG} format) and send a
remarkable number of photos (e.g.~the client is a photo shop) to
be double filtered and returned by the provider company; filters
must be applied in a pipeline scheme, i.e.~\emph{REDF} goes after
\emph{BWF}.

The structure of the system represented in
Fig.~\ref{fig:federated} corresponds to a \emph{federated system}.
It is defined as a system composed of components within different
administrative entities cooperating to provide a
service~\cite{federated}; this definition perfectly matches our
idea of SOA.

As a first example we consider the Classical semiring presented in
Sec.~\ref{sec:Dependqual}, therefore, in practice we show a crisp
constraint case. We suppose to have four variables
\texttt{outcomp}, \texttt{incomp}, \texttt{bwbyte} and
\texttt{redbyte}, which respectively represent the size in bytes
of the photo at the beginning of the process, after applying the
black-and-white filter, the red filter and after compressing the
obtained black-and-white photo. Since the client has a limited
memory space, it wants that the memory occupied by the photo does
not increase after the filtering and compressing process:

\[
\s{Memory}  \equiv \s{incomp} \leq \s{outcomp}
\]

The following three constraints represent the policies compiled
respectively by the staff of the \emph{BWF} module, the
\emph{REDF} module and \emph{COMPF} module. They state, following
their order, that applying the \emph{BWF} filter reduces the size
of the image,  applying the \emph{REDF} filter reduces the size of
the received black-and-white image and, at last, compressing the
image reduces its size.

\[
\s{BWFilter} \equiv \s{bwbyte} \leq \s{outcomp}
\]

\[
\s{REDFilter}\equiv  \s{redbyte} \leq     \s{bwbyte}
\]

\[
\s{Compression}\equiv  \s{incomp} \leq     \s{redbyte}
\]

The integration of the three policies (i.e.~soft constraints)
describes

\[
\s{Imp1} \equiv \s{BWFilter} \otimes \s{RedFilter} \otimes
\s{Compression}
\]

Integrity is ensured  in this system since \s{Imp1} ensures the
high-level requirement \s{Memory}.
\[
\s{Imp1}_{\Downarrow \{{\sf incomp},{\sf outcomp}\}} \sqsubseteq
\s{Memory}
\]

We are unconcerned about the possible values of the `internal'
variables \texttt{bwbyte} and \texttt{redbyte} and thus the
constraint relation $\s{Imp1}_{\Downarrow \{{\sf incomp},{\sf
outcomp}\}}$ describes the constraints in $\s{Imp1}$ that exist
between variables {\sf incomp} and {\sf outcomp}\}. By definition,
the above equation defines that all of the possible solutions of
$\s{Imp1}_{\Downarrow \{{\sf incomp},{\sf outcomp}\}}$ are
solutions of $\s{Memory}$, that is, for any assignment $\eta$ of
variables then
\[
\s{Imp1}_{\Downarrow \{{\sf incomp},{\sf outcomp}\}}~{\eta}
 \leq_{S} \s{Memory}~{\eta}
\]

\begin{definition}
We say that the requirement $S$ {\it locally refines \/}
requirement $R$ through the interface described by the set of
variables $V$ iff $S_{\Downarrow V} \sqsubseteq R_{\Downarrow V}$.
\end{definition}

Continuing the example in Fig.~\ref{fig:federated},  we assume
that the application system will behave reliably and uphold
\s{BWFilter} and \s{Compression}. Let us suppose instead that it
is not reasonable to assume that \emph{REDF} will always act
reliably, for example because the software of the red filter has a
small bug when the size of the photo is $666$Kbyte. In practice,
\emph{REDF} could take on any behavior:
\begin{eqnarray*}
\so{RedFilter} &\equiv& (\s{redbyte} \leq     \s{bwbyte} \vee
\s{redbyte}
> \s{bwbyte}) = true
\\
\s{Imp2} &\equiv& \s{BWFilter} \otimes \so{RedFilter} \otimes
\s{Compression}
\end{eqnarray*}
\s{Imp2} is a more realistic representation of the actual
filtering process. It more accurately reflects the reliability of
its infrastructure than the previous design \s{Imp1}. However,
since \s{redbyte} is no longer constrained it can take on any
value, and therefore, \s{incomp} is unconstrained and we have
\[
\s{Imp2}_{\Downarrow \{{\sf incomp},{\sf outcomp}\}}
\not\sqsubseteq \s{Memory}
\]
that is, the implementation of the system is not sufficiently
robust to be able to deal with internal failures  in a safe way
and uphold the memory probity requirement.

In~\cite{foley1,foley2} the author argues that this notion of
dependability may be viewed as a class of refinement whereby the
nature of the reliability of the system is explicitly specified.

\begin{definition}
(Dependability and Constraints~\cite{simon}) If $R$ gives
requirements for an enterprise and $S$ is its proposed
implementation, including details about the nature of the
reliability of its infrastructure, then $S$ is as {\it dependably
safe\/} as $R$ at interface that is described by the set of
variables $E$ if and only if $S_{\Downarrow E} \sqsubseteq
R_{\Downarrow E}$.
\end{definition}

\paragraph{\textbf{Quantitative analysis.}} When a quantitative analysis of the system is required, then it
is necessary to represent these properties  using soft
constraints. This can be done by simply considering a different
semiring (see Sec.~\ref{sec:Dependqual}), while the same
considerations provided for the previous example with crisp
constraints (by using the Classical semiring) still hold.

With a quantitative analysis, now consider that we aim not only to
have a correct implementation, but, if possible, to have the
``best'' possible implementation. We keep the photo editing
example provided in Fig.~\ref{fig:federated}, but we now represent
the fact that constraints describe the reliability percentage,
intended as the probability that a module will perform its
intended function. For example, the following (probabilistic) soft
constraint $c_1:\{\s{outcomp},\s{bwbyte}\} \rightarrow \mathbb{N}
\rightarrow [0,1]$ shows how the compression reliability performed
in \s{BWFilter} is linked to the initial and final number of bytes
of the treated image:

\begin{figure}\center{
$c_1(\s{outcomp},\s{bwbyte}) =
\begin{cases}
1   & \text{if $\s{outcomp} \leq 1024Kb$},\\
0   & \text{if $\s{outcomp} > 4096Kb$},\\
1 - \frac{\s{outcomp}}{100 \; \; \cdot \; \; \s{bwbyte}} & \text{otherwise}.\\
\end{cases}$}
\label{fig:pconstraints}
\end{figure}

$c_1$ tells us that the compression does not work if the input
image is more than $4Mb$, while is completely reliable if is less
than $1Mb$. Otherwise, this probability depends on the compression
efficiency: more that the image size is reduced during the
compression, more that it is possible to experience some errors,
and the reliability consequently decreases. For example,
considering the definition of $c_1$, if the input image is
$4096Kb$ and compressed is $1024Kb$, then the probability
associated to this variable instantiation is $0.96$.

In the same way, we can define $c_2$ and $c_3$ that respectively
shows the reliability for the \s{REDFilter} and \s{Compression}
modules. Their composition $\s{Imp3}= c_1 \otimes c_2 \otimes c_3$
represents the global reliability of the system. If
$\s{Memory}_{Prob}$ is the soft constraint representing the
minimum reliability that the system must provide (e.g.~$\s{Memory}_{Prob}$ is expressed by a client of the photo editing
system), then if

\[
\s{Memory} \sqsubseteq \s{Imp3}
\]

we are sure that the reliability requirements are entailed by our
system. Moreover, by exploiting the notion of best level of
consistency (see the \emph{blevel} in Sec.~\ref{sec:bgcon}), we
can find the best (i.e.~the most reliable) implementation among
those possible.

At last, notice also that the projection operator (i.e.~the
$\Downarrow$ operator explained in Sec.~\ref{sec:bgcon}) can be
used to model a sort of function declaration to the ``outside
world'': soft constraints represent the internal implementation of
the service, while projecting  over some variables leads to the
interface of the service, that is what is visible to the other
software components.

\section{A Nonmonotonic sccp Language for the Negotiation}\label{sec:nego}

In this Section we present a formal language based on soft
constraints~\cite{vodca}; the language is tied to the monitoring
of QoS aspects, as shown in Subsec.~\ref{sec:nonmonoex}. Given a
soft constraint system as defined in Sec.~\ref{sec:bgcon} and any
related constraint $c$, the syntax of agents in \emph{nmsccp} is
given in Fig.~\ref{tab:nmscc}. $P$ is the class of programs, $F$
is the class of sequences of procedure declarations (or clauses),
$A$ is the class of agents, $c$ ranges over constraints, $X$ is a
set of variables and $Y$ is a tuple of variables.

\begin{figure}[h]
\begin{center}
\scalebox{1}{
\begin{tabular}{ll}
$P ::= \: \:$ &  $F.A$\\
$F ::= \: \:$ & $p(Y)::A \mid F.F$\\
$A ::= \: \:$ & $success \mid tell(c) \rightarrowtail A \mid
retract(c) \rightarrowtail A  \mid update_X(c) \rightarrowtail A
\mid E \mid A \| A \mid \exists x.A \mid
p(Y)$\\
$E ::= \: \:$ & $ask(c) \rightarrowtail A  \mid nask(c)
\rightarrowtail A \mid E+E$
\end{tabular}
}
\end{center}
\caption{Syntax of the nmsccp language.} \label{tab:nmscc}
\end{figure}

The $\rightarrowtail$ is a generic checked transition used by
several actions of the language. Therefore, to simplify the rules
in Fig.~\ref{tab:transitionnmscc} we define a function
$check_{\rightarrowtail}: \sigma \rightarrow \{true, false\}$
(where $\sigma \in \C$), that, parametrized with one of the four
possible instances of $\rightarrowtail$ ({\textbf{C1}-{\textbf{C4}
in Fig.~\ref{tab:check}), returns {\em true} if the conditions
defined by the specific instance of $\rightarrowtail$ are
satisfied, or {\em false} otherwise. The conditions between
parentheses in Fig.~\ref{tab:check} claim that the lower threshold
of the interval clearly cannot  be ``better'' than the upper one,
otherwise the condition is intrinsically wrong.

In Fig.~\ref{tab:check} {\textbf{C1} checks if the
$\alpha$-consistency of the problem is between $a_1$ and $a_2$. In
words, {\textbf{C1} states that we need at least a solution as
good as $a_1$ entailed by the current store, but no solution
better than $a_2$; therefore, we are sure that some solutions
satisfy our needs, and none of these solutions is ``too good''.

\begin{figure}[h]
  \scalebox{0.89}{
  \begin{minipage}{0.5\linewidth}
    \begin{center}
\begin{tabular}{llll}


&\mbox{  }&\mbox{   } &\mbox{   }
\\
\mbox{\textbf{C1:} $\rightarrowtail=\rightarrow_{a_1}^{a_2}$}&
\mbox{\;\;$check(\sigma)_{\rightarrowtail}= true \:\text{if} $}&
{\;$
\begin{cases}
\sigma\Downarrow_{\emptyset} \not>_S a_2
\\
\sigma\Downarrow_{\emptyset} \not<_S a_1
\end{cases}$}&
\\

&\mbox{(with $a_1 \not> a_2$)}&\mbox{} &\mbox{   }
\\

\\
&\mbox{  }&\mbox{   } &\mbox{   }
\\
\mbox{\textbf{C2:} $\rightarrowtail=\rightarrow_{a_1}^{\phi_2}$}&
\mbox{\;\;$check(\sigma)_{\rightarrowtail}= true \:\text{if} $}&
{\; $
\begin{cases}
\sigma \not\sqsupset \phi_2
\\
\sigma\Downarrow_{\emptyset} \not<_S a_1
\end{cases}$}&
\\

&\mbox{(with $a_1 \not> \phi_2\Downarrow_{\emptyset}$)}&\mbox{}
&\mbox{ }
\\
&\mbox{   }&\mbox{   } &\mbox{   }
\\

\end{tabular}
    \end{center}
  \end{minipage}
   \hspace{1cm}
  \begin{minipage}{0.5\linewidth}
    \begin{center}
    \begin{tabular}{llll}

&\mbox{  }&\mbox{   } &\mbox{   }
\\
\mbox{\textbf{C3:} $\rightarrowtail=\rightarrow_{\phi_1}^{a_2}$}&
\mbox{\;\;$check(\sigma)_{\rightarrowtail}= true \:\text{if} $}&
{\; $
\begin{cases}
\sigma\Downarrow_{\emptyset} \not>_S a_2
\\
\sigma \not\sqsubset \phi_1
\end{cases}$}&
\\

&\mbox{(with $\phi_1\Downarrow_{\emptyset} \not> a_2$)}&\mbox{}
&\mbox{ }
\\

\\
&\mbox{  }&\mbox{   } &\mbox{   }
\\
\mbox{\textbf{C4:}
$\rightarrowtail=\rightarrow_{\phi_1}^{\phi_2}$}&
\mbox{\;\;$check(\sigma)_{\rightarrowtail}= true \:\text{if} $}&
{\; $
\begin{cases}
\sigma \not\sqsupset \phi_2
\\
\sigma \not\sqsubset \phi_1
\end{cases}$}&
\\

&\mbox{(with $\phi_1 \not\sqsupset \phi_2$)}&\mbox{} &\mbox{   }
\\

&\mbox{   }&\mbox{   } &\mbox{   }
\\
\end{tabular}
    \end{center}
  \end{minipage} }
\hspace{4cm}\footnotesize{ \textit{Otherwise, within the same
conditions in parentheses,
$check(\sigma)_{\rightarrowtail}=false$}} \caption{Definition of
the {\em check} function for each of the four checked
transitions.} \label{tab:check}
\end{figure}

To give an operational semantics to our language we need to
describe an appropriate transition system \mbox{$\langle \Gamma,
T, \rightarrow \rangle$}, where $\Gamma$ is a set of possible
configurations, $T \subseteq \Gamma $ is the set of {\em terminal}
configurations and $\rightarrow \subseteq \Gamma \times \Gamma$ is
a binary relation between configurations. The set of
configurations is $\Gamma = \{\langle A, \sigma\rangle\}$, where
$\sigma \in \mathcal{C}$ while the set of terminal configurations
is instead  $T=\{\langle success, \sigma \rangle\}$. The
transition rules for the {\em nmsccp} language are defined in
Fig.~\ref{tab:transitionnmscc}.

\begin{figure}
  \scalebox{0.89}{

  \begin{minipage}{0.5\linewidth}
    \begin{center}
\begin{tabular}{llll}

&\mbox{   }&\mbox{   } &\mbox{   }
\\
\mbox{\bf R1}& $\frac {\displaystyle check (\sigma \otimes
c)_{\rightarrowtail}}{\displaystyle
\begin{array}{l}
\la \hbox{\tell}(c) \rightarrowtail A, \sigma \ra \rrarrow \la A,
\sigma \otimes c\ra
\end{array}}$\ \ \ & \bf{Tell}&
\\

&\mbox{   }&\mbox{   } &\mbox{   }
\\
\mbox{\bf R2}& $\frac {\displaystyle \sigma \ent c \ \ \ \ \
check(\sigma)_{\rightarrowtail}}{\displaystyle
\begin{array}{l}
\la \hbox{\ask}(c) \rightarrowtail A, \sigma \ra \rrarrow \la A,
\sigma \ra
\end{array}}$\ \ \ & \bf{Ask}&
\\

&\mbox{   }&\mbox{   } &\mbox{   }
\\
\mbox{\bf R3}& $\frac {\displaystyle \la A,\sigma \ra \rrarrow \la
A', \sigma'   \ra} {\displaystyle
\begin{array}{l}
\la A\parallel B, \sigma \ra\rrarrow \la A'\parallel B, \sigma'
\ra
\\
\la B\parallel A, \sigma \ra\rrarrow \la B\parallel A', \sigma'
\ra
\end{array}}$& \bf{Parall1}&
\\

&\mbox{   }&\mbox{   }&\mbox{   }
\\
\mbox{\bf R4}& $\frac {\displaystyle \; \; \; \la A,\sigma\ra
\rrarrow \la success, \sigma'\ra \; \; \;} {\displaystyle
\begin{array}{l}
\la A\parallel B, \sigma \ra\rrarrow \la B, \sigma' \ra
\\
\la B\parallel A, \sigma \ra\rrarrow \la B, \sigma' \ra
\end{array}}$&
 \bf{Parall2}&
\\
&\mbox{   }&\mbox{   }&
\\

\mbox{\bf R5}& $\frac {\displaystyle \la  E_j,\sigma\ra \rrarrow
\la A_j,\sigma' \ra\ \ \ \ \ \ j\in [1,n]} {\displaystyle
\begin{array}{l}
\la \Sigma_{i=1}^{n}E_i , \sigma \ra\rrarrow \la A_j,\sigma'\ra
\end{array}}$ & \bf{Nondet}&
\\

\end{tabular}
    \end{center}
  \end{minipage}
   \hspace{1cm}
  \begin{minipage}{0.5\linewidth}
    \begin{center}
    \begin{tabular}{llll}

&\mbox{   }& \mbox{   }
\\
\mbox{\bf R6}& $\frac {\displaystyle \sigma \not\ent c \ \ \ \ \ \
check(\sigma)_{\rightarrowtail}} {\displaystyle\la nask(c)
\rightarrowtail A, \sigma \ra\rrarrow\la A, \sigma \ra}$
&\bf{Nask} &
\\

&\mbox{   }&\mbox{   } &\mbox{   }
\\
\mbox{\bf R7}& $\frac {\displaystyle  \sigma \sqsubseteq c  \ \ \
\ \ \sigma'= \sigma \, \odiv \, c \ \ \ \ \
check(\sigma')_{\rightarrowtail}} {\displaystyle\la retract(c)
\rightarrowtail A, \sigma \ra\rrarrow\la A, \sigma' \ra}$
&\bf{Retract} &
\\

&\mbox{   }&\mbox{   }&
\\
\mbox{\bf R8}& $\frac {\displaystyle \sigma' =
(\sigma\Downarrow_{(V \backslash X)})\otimes c \ \ \ \ \
check(\sigma')_{\rightarrowtail}} {\displaystyle\la update_X(c)
\rightarrowtail A, \sigma \ra\rrarrow\la A, \sigma' \ra}$
&\bf{Update} &
\\

&\mbox{   }&\mbox{   }&
\\
\mbox{\bf R9}& $\frac {\displaystyle \la A[x/y],  \sigma
\ra\rrarrow\la B, \sigma' \ra} {\displaystyle\la \exists x.A,
\sigma \ra\rrarrow\la B, \sigma' \ra}  \text{ with $y$ {\em
fresh}}$
&\bf{Hide} &\\
&\mbox{   }&\mbox{   }&
\\
\mbox{\bf R10}& $\frac {\displaystyle \la A,\sigma\ra\rrarrow\la
B, \sigma' \ra} {\displaystyle\la p(Y),\sigma\ra\rrarrow\la B,
\sigma'\ra} \ \ \ \ {\it p(Y) :: A \in F}$
&\bf{P-call}&\\
\\
\end{tabular}
    \end{center}
  \end{minipage} }

\caption{The transition system for {\em nmsccp}.}
\label{tab:transitionnmscc}
\end{figure}

In the following we provide a description of the transition rules
in Fig.~\ref{tab:transitionnmscc}.  For further details, please
refer to~\cite{vodca}. In the {\bf Tell} rule ({\bf R1}), if the
store $\sigma \otimes c$ satisfies the conditions of the specific
$\rightarrowtail$ transition of Fig.~\ref{tab:check}, then the
agent evolves to the new agent $A$ over the store $\sigma \otimes
c$. Therefore the constraint $c$ is added to the store $\sigma$.
The conditions are checked on the (possible) next-step store: i.e.~$check(\sigma')_{\rightarrowtail}$. To apply the {\bf Ask} rule
({\bf R2}), we need to check if the current store $\sigma$ entails
the constraint $c$ and also if the current store is consistent
with respect to the lower and upper thresholds defined by the
specific $\rightarrowtail$ transition arrow: i.e.~if
$check(\sigma)_{\rightarrowtail}$ is true.

{\bf Parallelism and nondeterminism}: the composition operators
$+$ and $\|$ respectively model nondeterminism and parallelism. A
parallel agent (rules {\bf R3} and {\bf R4}) will succeed when
both agents succeed. This operator is modelled in terms of {\em
interleaving} (as in the classical {\itshape ccp}): each time, the
agent $A\parallel B$ can execute only one between the initial
enabled actions of $A$ and $B$ ({\bf R3}); a parallel agent will
succeed if all the composing agents succeed ({\bf R4}). The
nondeterministic rule {\bf R5} chooses one of the agents whose
guard succeeds, and clearly gives rise to global nondeterminism.
The {\bf Nask} rule is needed to infer the absence of a statement
whenever it cannot be derived from the current state: the
semantics in {\bf R6} shows that the rule is enabled when the
consistency interval satisfies the current store (as for the {\em
ask}), and $c$ is not entailed by the store: i.e.~$\sigma
\not\sqsubseteq c$. {\bf Retract}: with {\bf R7} we are able to
``remove'' the constraint $c$ from the store $\sigma$, using the
$\odiv$ constraint division function defined in
Sec.~\ref{sec:bgcon}. According to {\bf R7}, we require that the
constraint $c$ is entailed by the store, i.e.~$\sigma \sqsubseteq
c$. The semantics of {\bf Update} rule ({\bf R8})~\cite{NMCC2}
resembles the assignment operation in imperative programming
languages: given an $update_X(c)$, for every $x \in X$ it removes
the influence over $x$ of each constraint in which $x$ is
involved, and finally a new constraint $c$ is added to the store.
To remove the information concerning all $x\in X$, we project (see
Sec.~\ref{sec:bgcon}) the current store on $V\backslash X$, where
$V$ is the set of all the variables of the problem and $X$ is a
parameter of the rule (projecting means eliminating some
variables). At last, the levels of consistency are checked on the
obtained store, i.e.~$check(\sigma')_{\rightarrowtail}$. Notice
that all the removals and the constraint addition are
transactional, since are executed in the same rule. {\bf Hidden
variables}: the semantics of the existential quantifier in {\bf
R9} can be described by using the notion of {\em freshness} of the
new variable added to the store~\cite{vodca}. {\bf Procedure
calls}: the semantics of the procedure call ({\bf R10}) has
already been defined in~\cite{scc}: the notion of diagonal
constraints (as defined in Sec.~\ref{sec:bgcon}) is used to model
parameter passing.

\subsection{Example}\label{sec:nonmonoex}

One  application of the {\em nmsccp} language is to model generic
entities negotiating a formal agreement, i.e.~a
SLA~\cite{federated,wslaframework}, where the level of service is
formally defined. The main task consists in accomplishing the
requests of all the agents by satisfying their QoS requirements.
Considering the fuzzy negotiation in Fig.~\ref{fig:fuzzy}
(\emph{Fuzzy} semiring: $\langle [0,1], max, min, 0 ,1 \rangle$)
both a provider and a client (offering and acquiring a web
service, for example) can add their request to the store $\sigma$
(respectively $tell(c_p)$ and $tell(c_c)$): the thick line
represents the consistency of $\sigma$ after the composition (i.e.~$min$), and the \emph{blevel} of this SCSP (see
Sec.~\ref{sec:bgcon}) is the \emph{max}, where both requests
intersects (i.e.~in $0.5$).

\begin{figure}
\centering
    \includegraphics[scale=0.75]{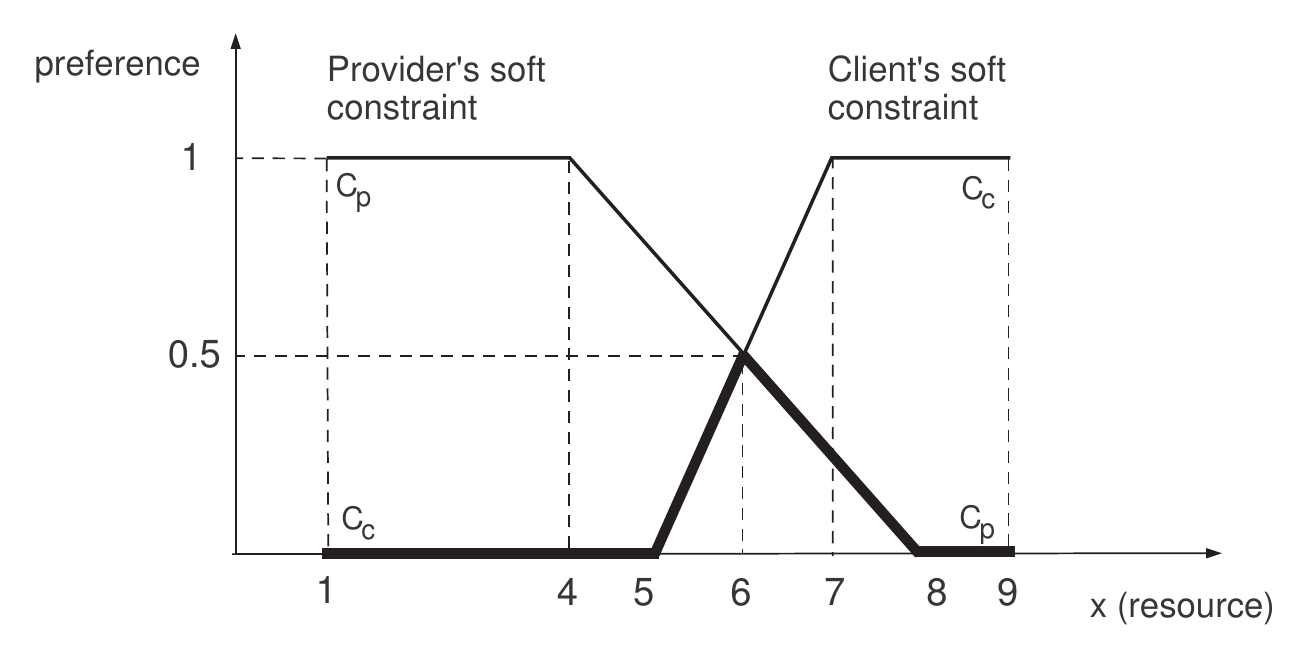}
\caption{The graphical interpretation of a fuzzy
agreement.}\label{fig:fuzzy}
\end{figure}

We present three short examples to suggest possible negotiation
scenarios. We suppose there are two distinct companies (e.g.~providers $P_1$ and $P_2$) that want to merge their services in a
sort of pipeline, in order to offer to their clients a single
structured service: e.g.~$P_1$ completes the functionalities of
$P_2$. This example models the \emph{cross-domain} management of
services proposed in~\cite{federated}. The variable $x$ represents
the global number of failures they can sustain during the service
provision, while the preference models the number of hours (or a
money cost in hundreds of euros) needed to manage them and recover
from them. The preference interval on transition arrows models the
fact that both $P_1$ and $P_2$ explicitly want to spend some time
to manage the failures (the upper bound in Fig.~\ref{tab:check}),
but not so much time (lower bound in Fig.~\ref{tab:check}). We
will use the  \emph{Weighted} semiring and the soft constraints
given in Fig.~\ref{fig:fuzzyexample}. Even if the examples are
based on a single criterion (i.e.~the number of hours) for sake of
simplicity, they can be extended to the multicriteria case, where
the preference is expressed as a tuple of incomparable criteria.

\begin{example}[Tell and negotiation]\label{ex:1}
$P_1$ and $P_2$ both want to present their policy (respectively
represented by $c_4$ and $c_3$) to the other party and to find a
shared agreement on the service (i.e.~a SLA). Their agent
description is: $P_1 \equiv \langle tell(c_4)
\rightarrow_{\infty}^{0} tell(s_{p2}) \rightarrow_{\infty}^{0}
ask(s_{p1}) \rightarrow_{10}^{2} success \rangle ||
 \langle tell(c_3) \rightarrow_{\infty}^{0} tell(s_{p1})
\rightarrow_{\infty}^{0} ask(s_{p2}) \rightarrow_{4}^{1} success
\rangle \equiv P_2$, executed in the store with empty support
(i.e.~$\bar{0}$). Variables $s_{p1}$ and $s_{p2}$ are used only
for synchronization and thus will be ignored in the following
considerations (e.g.~replaced by the $SYNCHRO_i$ agents in
Ex.~\ref{ex:2}). The final store (the merge of the two policies)
is $\sigma = (c_4 \otimes c_3) \equiv 2x + x + 5$, and since
$\sigma\Downarrow_{\emptyset} = 5$ is not included in the last
preference interval of $P_2$ (between $1$ and $4$), $P_2$ does not
succeed and a shared agreement cannot be found. The practical
reason is that the failure management systems of $P_1$ need at
least $5$ hours (i.e.~$c_4 = x + 5$) even if no failures happen
(i.e.~$x = 0$). Notice that the last interval of $P_2$ requires
that at least $1$ hour is spent to check failures.
\end{example}

\begin{figure}
\begin{align*}
c_1:(\{x\} \rightarrow  \mathbb{N}) \rightarrow \mathbb{R}^+
 \; \; \text{ s.t. } c_1(x) = x + 3 \hspace{1.1cm} c_2:(\{y\} \rightarrow \mathbb{N}) \rightarrow
\mathbb{R}^+ \, \, \text{s.t.} \; c_2(y) = y+1
\end{align*}
\begin{center}
$\ \, c_3:(\{x\} \rightarrow \mathbb{N}) \rightarrow \mathbb{R}^+ \; \;
\text{ s.t. }  c_3(x) = 2x$ \hspace{1.25cm} $c_4:(\{x\} \rightarrow
\mathbb{N}) \rightarrow \mathbb{R}^+ \; \text{ s.t. } c_4(x) =
x+5$\end{center}\vspace{-0.4cm} \caption{Four Weighted soft
constraints.}\label{fig:fuzzyexample}
\end{figure}

\begin{example}[Retract]\label{ex:2}
After some time (still considering Ex.~\ref{ex:1}), suppose that
$P_1$ wants to relax the store, because its policy is changed:
this change can be performed from an interactive console or by
embedding timing mechanisms in the language as explained
in~\cite{coordtimed}. The removal is accomplished by retracting
$c_1$, which means that $P_1$ has improved its failure management
systems. Notice that $c_1$ has not ever been added to the store
before, so this retraction behaves as a relaxation; partial
removal is clearly important in a negotiation process. $P_1 \equiv
\langle tell(c_4) \rightarrow_{\infty}^{0} SYNCHRO_{P1}
\rightarrow_{10}^{2} retract(c_1) \rightarrow_{10}^{2} success
\rangle ||
 \langle tell(c_3) \rightarrow_{\infty}^{0} SYNCHRO_{P2} \rightarrow_{4}^{1} success
\rangle \equiv P_2$ is executed in $\bar{0}$. The final store is
$\sigma = c_4 \otimes c_3 \, \odiv\, c_1 \equiv 2x + 2$, and since
$\sigma\Downarrow_{\emptyset} = 2$, both $P_1$ and $P_2$ now
succeed (it is included in both intervals).
\end{example}

\begin{example}[Update]\label{ex:4}The \emph{update} can instead be  used for substantial changes of
the policy: for example, suppose that $P_1 \equiv \langle
tell(c_1) \rightarrow_{\infty}^{0} update_{\{x\}}(c_2)
\rightarrow_{\infty}^{0} success, \bar{0}\rangle$. This agent
succeeds in the store $\bar{0} \otimes c_1 \Downarrow_{(V
\backslash \{x\})} \otimes c_2$, where $c_1 \Downarrow_{(V
\backslash \{x\})} = \bar{3}$ and $\bar{3} \otimes c_2 \equiv y +
4$ (i.e.~the polynomial describing the final store). Therefore,
the first policy based on the number of failures (i.e.~$c_1$) is
updated such that $x$ is ``refreshed'' and the newly added policy
(i.e.~$c_2$) depends only on the  number $y$ of system reboots.
The consistency level of the store (i.e.~the number of hours) now
depends only on the $y$ variable of the SCSP. Notice that the
$\bar{3}$ component of the final store derives from the ``old''
$c_1$, meaning that some fixed management delays are included also
in this new policy.
\end{example}

\section{Related Work}\label{sec:related}

There exist already several proposals for languages which allow to
specify (Web) services and their composition, at different levels
of abstractions, ranging from description languages such as
\emph{WSDL} (which allows to describe services essentially as
collections of ports), to orchestration languages (\emph{XLANG},
\emph{WSFL} and \emph{WS-BPEL}) and choreography languages
(\emph{WS-CDL} and \emph{BPEL4Chor}) which allow to define
composition of services either in terms of a centralized
meta-service (the orchestrator) or by considering the reciprocal
interactions (the choreography) among the different services
(without centralization). There exist also some specific
proposals~\cite{laneve,zavattaro} for describing contracts and
their relevant properties. However a general, established theory
of contracts is still missing. Furthermore most languages do not
take into account SLAs, that is aspects of contracts such as cost,
performance or availability, which are related to QoS.

Other papers have been proposed in order to study dependability
aspects in SOAs, for example by using the  \emph{Architecture
Analysis and Design Language} (AADL). In \cite{depbg1} the authors
purpose a modeling framework allowing the generation of
dependability-oriented analytical models from AADL models, to
facilitate the evaluation of dependability measures, such as
reliability or availability. The AADL dependability model is
transformed into a  \emph{Generalized Stochastic Petri Net} (GSPN)
by applying model transformation rules.

The frameworks presented in this paper can join the other formal
approaches for architectural notations: Graph-Based, Logic-Based
and Process Algebraic approaches~\cite{fapproaches}.  A formal
foundation underlying is definitely important, since, for example,
UML alone can offer a number of alternatives for representing
architectures: therefore, this lack of precision can lead to a
semantic misunderstanding of the described architectural
model~\cite{misund}. Compared to other formal
methods~\cite{fapproaches}, constraints are very expressive and
close to the human way of describing properties and relationships;
in addition, their solution techniques have a long and successful
history~\cite{bookrossi}. The qualitative/quantitative
architectural evaluation can be accomplished by considering
different semirings (see Sec~\ref{sec:bgcon}): our framework is
highly parametric and can consequently deal with different QoS
metrics, as long as they can be represented as semirings.

Other works have studied the problem of issuing requests to a
composition of web services as a crisp constraint-based
problem~\cite{laz:cp05}, but without optimizing non-functional
aspects of services, as we instead do with (semiring-based) soft
constraints. For a more precise survey on the architectural
description of dependable software systems, please refer to
\cite{surveydep}. The most direct comparison for {\em nmsccp} in
Sec.~\ref{sec:nego} is with the work in~\cite{cc-pi}, in which
soft constraints are combined with a name-passing calculus. The
most important difference is that in {\em nmsccp} we do not have
the concept of constraint token and it is possible to remove every
$c$ that is entailed by the store, even if $c$ is syntactically
different from all the $c$ previously
added. 

\section{Conclusions and Future Work}\label{sec:onclusions}
We have proved that soft constraints and their related operators
(e.g.~$\otimes$, $\odiv$, $\Downarrow$ in Sec.~\ref{sec:bgcon})
can model and manage the composition of services in SOAs by taking
in account QoS metrics. The key idea is that constraint
relationships model the implementation of a service component
(described as a policy), while the ``softness'' (i.e.~the
preference value associated with the soft constraint) represents
one or more QoS measures, as reliability, availability and so on
(see Sec.~\ref{sec:Dependqual}). In this way, the composition of
services can be monitored and checked, and  the best quality
result can be found on this integration. It may also be desirable
to describe constraints and capabilities (also, ``policy'')
regarding security: a web service specification could require
that, for example, ``you MUST use HTTP Authentication and MAY use
GZIP compression''.

Two different but very close contributions are collected in this
work. The first contribution in Sec.~\ref{sec:conn} is that the
use of soft constraints permits one to perform a quantitative
analysis of system integrity. The second contribution, explained
in Sec.~\ref{sec:nego}, proposes the use of a formal language
based on soft constraints in order to model the composition of
different service components while monitoring QoS aspects at the
same time.

All the models and techniques presented in this work can be
implemented and integrated together in a suite of tools, in order
to manage and monitor QoS while building  SOAs. To accomplish this
task, we could  extend an existent solver such as
Gecode~\cite{gecode}, which is an open, free, portable,
accessible, and efficient environment for developing
constraint-based systems and applications. The main results would
be the development of a SOA query engine, that would use the
constraint satisfaction solver to select which available service
will satisfy a given query. It would also look for complex
services by composing together simpler service interfaces.

\bibliographystyle{eptcs}
\bibliography{references}

\end{document}